%% file: main.tex
\documentclass[10pt,twocolumn,letterpaper]{article}

\usepackage{iccv}              % To produce the CAMERA-READY version
\usepackage[table,xcdraw]{xcolor} % For coloring
\usepackage{colortbl} % For more advanced table formatting
\usepackage{graphicx} % For resizing
\usepackage{booktabs} % For better table rules

\definecolor{iccvblue}{rgb}{0.21,0.49,0.74}
\usepackage[pagebackref,breaklinks,colorlinks,allcolors=iccvblue]{hyperref}
\def\paperID{11198} % *** Enter the Paper ID here
\def\confName{ICCV}
\def\confYear{2025}

\newcommand{\fyq}[1]{\textcolor{black}{#1}}

\title{AHMAD: \underline{A}daptive \underline{H}ybrid \underline{M}ulti-task Vision Learning with \underline{A}ssisted \underline{D}istillation}
\author{Mohammad Mahdi\thanks{Corresponding author} \quad Nedyalko Prisadnikov \quad Yuqian Fu \quad Carmelo Scribano \\ Danda Pani Paudel \quad Luc Van Gool \\
INSAIT, Sofia University “St. Kliment Ohridski” \\
\texttt{\{firstname.lastname\}@insait.ai}}
\begin{document}
\maketitle
\input{sec/0_abstract}    
\input{sec/1_intro}

\input{sec/2_relatedw}
\input{sec/3_method}

\input{sec/4_exps}

\input{sec/5_conc}
{
    \small
    \bibliographystyle{ieeenat_fullname}
    \bibliography{main}
}

\clearpage

\end{document}

%% file: sec/0_abstract.tex
\begin{abstract}
Generalist multitasking vision models aim to unify multiple vision tasks within a single framework, enabling more efficient and versatile learning. However, handling diverse vision tasks—spanning dense and sparse predictions—remains challenging due to their inherently varying output structures.
In this paper, we propose AHMAD, a simple yet effective framework for generalist multitask learning that integrates different key vision tasks: semantic segmentation, instance segmentation, depth estimation, keypoint detection, and object detection. 
Our approach incorporates these five tasks into a unified structure -- a shared encoder-decoder with several lightweight task-specific projectors. Under the multitask learning paradigm, we observed a complementary performance gain, achieving a state-of-the-art \textbf{PQ} of \textbf{53.1} and an \textbf{mIoU} of \textbf{66.5} for $\texttt{COCO-val}$ panoptic and semantic segmentation, respectively.
Additionally, for top-down keypoint detection, which typically incurs high computational overhead due to multiple forward passes, we introduce a knowledge distillation-based method that enables a single forward pass over the entire image, greatly improving efficiency.
Ultimately, our model delivers a lightweight yet effective generalist multitask learning framework, demonstrating strong performance across five vision tasks. 

\end{abstract}

%% file: sec/1_intro.tex
\section{Introduction}
\label{sec:intro}
\begin{figure}[t]
  % \hspace{-9px} % Adjust the value as needed 
  \includegraphics[width=0.5\textwidth, height=0.6\linewidth]{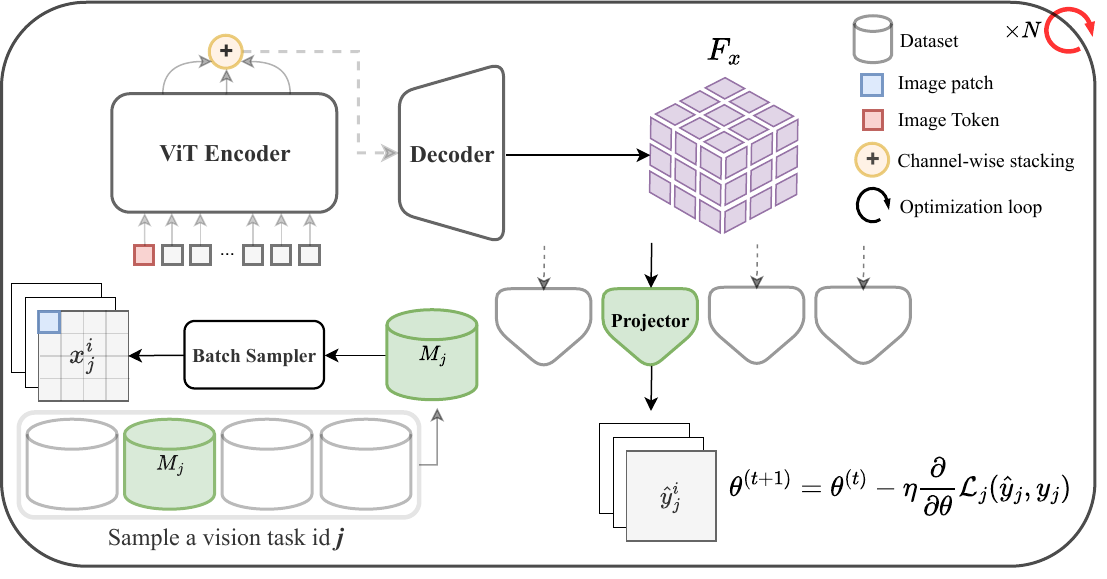}
   \caption{\textbf{Our Proposed Training Framework}. We leverage a powerful vision transformer encoder combined with a CNN-based decoder to generate a rich feature map, which contains representations useful for various tasks. Task-specific projectors are then applied to perform different tasks. 
   %The training loop of our model consists of $N=\frac{\max_j \left( size(M_j)\right)}{batch\_size}$ optimization steps, each corresponding to a randomly selected task \(j\).
   }
%    \caption{\textbf{Our proposed training framework}. A single training loop of our model consists of $N=\frac{\max_j \left( \lvert M_j \rvert \right)}{bs}$ optimization steps, each over a randomly sampled task \(j\). \(bs\) is batch size, \(\lvert.\rvert\) denotes cardinality, and the \( \oplus \) operation performs concatenation over the outputs of the backbone's intermediate layers.
% }
\vspace{-0.1in}
   \label{fig:fig1}
\end{figure}

%p1: 
Generalist models aim to solve multiple tasks within a single unified framework, eliminating the need for separate task-specific models. Given the inherently multitask nature, technically, building generalist models is closely aligned with multi-task learning, sharing the goal of optimizing a model to handle diverse tasks efficiently.

In the NLP domain, generalist multitask learning has achieved remarkable success, with models such as ~\cite{achiam2023gpt, guo2025deepseek}, demonstrating strong performance across diverse language tasks. This success is largely attributed to the shared nature of text representations, where different language tasks can be formulated within a common sequence-to-sequence or auto-regressive framework. However, developing generalist multitask learning in vision remains is significantly more challenging due to the inherent diversity in vision tasks. Vision tasks span a broad range of dense and sparse predictions, each requiring distinct output structures and optimization objectives. For example, semantic segmentation operates at the pixel level, depth estimation requires continuous value regression, while object detection involves structured bounding box predictions. These fundamental differences make it difficult to design a generalist framework that seamlessly accommodates all vision tasks within a single model.

% \textcolor{green}{Generalist} vision models are encoder-decoder architectures designed to address a broad range of vision problems within a unified framework. These models leverage a shared encoder to extract rich visual representations, which are then processed by a task-agnostic decoder to generate feature-rich solutions across multiple domains. 

% Multi-tasking in this context refers to making predictions on diverse tasks by projecting decoder features onto task-specific outputs, enabling a single model to perform various vision tasks efficiently. This approach enhances adaptability, reduces redundancy, and fosters knowledge sharing across tasks.

% The key challenge in generalist multitasking models is managing task diversity. Vision tasks span a wide range, including both sparse and dense predictions, which complicates their alignment within a unified framework. This difficulty arises primarily from their heterogeneous output structures. For instance, while semantic segmentation requires per-pixel predictions, object detection involves outputs that are not tied to individual pixels but rather to object instances.

To tackle this, some studies focus on task output homogenization, where various task outputs are manually encoded into a shared format through hand-crafted methods. In this context, Painter~\cite{wang2023images} approaches many vision tasks as image inpainting problems, encoding task outputs in the RGB space. Pix2Seq-D~\cite{chen2023generalist} also leverages the Bit Diffusion model~\cite{chen2022analog} to learn task-specific outputs by converting them into per-pixel representations across separate channels. Alternatively, other approaches add components to serialize both the input image and task outputs, treating the problem as a next-token prediction task. For instance, Unified-IO~\cite{lu2022unified} employs a VQ-GAN~\cite{esser2021taming} to serialize dense tasks and adds special tokens for sparse tasks. In this paradigm, the generalist model is trained on the frozen tokens, and the outputs need to be further decoded. However, both these approaches come with trade-offs, as they introduce complexity through demanding pre- and post-processing, along with the need for sophisticated training techniques, such as VQ-GAN for serialization.

In generalist multitasking, it is also essential to unify different vision tasks under a common problem-solving methodology. Among these tasks, person keypoint detection often follows a top-down strategy, where individuals are cropped from the image, and keypoints are then detected for each cropped person. This approach has become the standard due to the effectiveness of person detectors. However, it requires running the model multiple times, once for each individual in the image, which goes against the core idea of generalist multitasking, where the goal is to process the entire image and perform multiple tasks in a single forward pass. This multiple-run process is especially expensive in generalist vision models, which rely on large, computationally intensive encoder backbones. Despite this, no significant effort has been made to address this issue while still leveraging the strengths of the top-down approach.

We propose a simple yet effective training framework, as in Fig.~\ref{fig:fig1}, namely AHMAD, for generalist multitasking across different vision tasks: Panoptic Segmentation (Semantic Segmentation (SS) and Instance Segmentation (IS)), Depth Estimation (DE), Keypoint Detection (KD), and Object Detection (OD), demonstrating a complementary boost for the majority of tasks. Additionally, we introduce a novel method to reduce the forward passes required for top-down keypoint detection, achieving a single run for the entire image using knowledge distillation techniques. 

Our contributions are as follows: 1) We present a simple multitask learning paradigm with no restrictions on task output shapes or the need for extra components, allowing the incorporation of each task's ideal single-task approach within a unified multitask structure. 2) We propose a novel distillation method to perform top-down keypoint detection with just a single forward pass for the entire image. 3) We demonstrate strong performance across various vision tasks, achieving a PQ of 53.1 and an mIoU of 66.5 for COCO panoptic and semantic segmentation among generalist models, while also delivering competitive results in keypoint detection and object detection.

%% file: sec/2_relatedw.tex
\section{Related Work}
\label{sec:relatedw}
\textbf{Vision Transformers.} Vision Transformers (ViTs)~\cite{khan2022transformers,ranftl2021vision} have emerged as powerful foundations for both multi-task and multimodal learning in computer vision~\cite{thung2018brief}. Swin Transformer~\cite{liu2021swin} introduces hierarchical feature by partitioning images into non-overlapping windows and computing self-attention within each. This enables efficient modeling of long-range dependencies through a shifting window mechanism. ViTs also excel in self-supervised learning~\cite{liu2019end, standley2020tasks, zhang2018overview}, acting as strong backbones for various vision tasks. Methods like DINO~\cite{caron2021emerging} and DINOv2~\cite{oquab2023dinov2} leverage unsupervised pretraining based on knowledge distillation on large unlabeled data, enabling them to learn transferable representations for different downstream tasks. Furthermore, ViTs are effective in multimodal learning~\cite{huang2021makes}, particularly in vision-language tasks~\cite{zhang2024vision} such as visual question answering~\cite{ soni2024earthdial}, where combining visual and textual data enhances overall comprehension.

\textbf{Segmentation and Object Detection.} Transformer-based models excel in dense tasks like segmentation~\cite{minaee2021image}. MaskFormer~\cite{cheng2021per} frames segmentation as hierarchical mask generation and classification, while Mask2Former~\cite{cheng2022masked} refines it with multi-scale masked attention. OneFormer~\cite{jain2023oneformer} extends this with task-conditioned training for handling diverse segmentation tasks. SAM~\cite{kirillov2023segment} generates high-quality object masks from both sparse and dense prompts, using a MAE-pretrained Vision Transformer~\cite{he2022masked} as the encoder and a prompt encoder for boxes, points, and text (encoded using CLIP~\cite{radford2021learning}), with support for dense mask prompts encoded by a CNN network. SAM2~\cite{ravi2024sam} builds on this with a promptable visual segmentation framework, incorporating a memory bank and attention mechanism for improved accuracy.  Transformers~\cite{vaswani2017attention} are also widely used for open-vocabulary object detection~\cite{zareian2021open,gu2021open,minderer2022simple,du2022learning,minderer2023scaling,cheng2024yolo}, leveraging autoregressive next-token prediction. Grounding DINO~\cite{liu2024grounding} detects objects from language inputs, with a Swin transformer for image encoding, BERT~\cite{devlin2019bert} for text, and deformable self-attention~\cite{xia2022vision} for feature enhancement. It also employs language-guided query selection and a cross-modality decoder for bounding box prediction. Grounding DINO 1.5~\cite{ren2024grounding} enhances this with Pro and Edge versions, offering a larger ViT-L backbone and cross-scale feature fusion for improved performance. Grounded SAM~\cite{ren2024grounded} merges Grounding DINO and SAM for both segmentation and detection, using BLIP~\cite{li2022blip,li2023blip} and RAM~\cite{zhang2024recognize} for vision-language understanding.

\textbf{Multitask Generalist Models.} Expanding beyond models that are strictly designed for segmentation and detection tasks, Painter~\cite{wang2023images} frames dense vision tasks as image inpainting, with outputs in the same RGB space as input images. The task prompt is encoded as a pair of images, and a Masked Vision Transformer is used for training. At inference, Painter offers flexible task adaptation through three types of prompts: random, searched, and learned. Unified-IO~\cite{lu2022unified,lu2024unified} introduced an encoder-decoder architecture, using stacked transformer layers to represent all inputs and outputs as discrete tokens. It utilizes VQ-GAN~\cite{esser2021taming} for image serialization and incorporates location tokens and sparse structures like bounding boxes as special tokens. However, the model has limitations in localization. Pix2Seq~\cite{chen2022unified} takes an image and task prompts to generate discrete tokens corresponding to desired outputs, while Pix2Seq-D~\cite{chen2023generalist} handles semantic and instance segmentation with two channels per task. GiT~\cite{wang2024git} divides the image into subregions, generating sparse responses for each, which are then merged. DINO-X~\cite{ren2024dino} extends the Grounding DINO encoder-decoder architecture for open-world understanding, supporting text, visual, and customized prompts. The DINO-X Pro version uses a pre-trained CLIP model to enhance performance across multimodal benchmarks, with visual prompts leveraging box and point formats for better segmentation and visual grounding tasks.

% We follow Nedyalko’s work~\cite{prisadnikov2024simple} for its flexibility in combining dense and sparse tasks. The approach uses a DinoV2 backbone  plus a CNN-based Decoder, with different heads for each task, including Semantic/Instance Segmentation and Depth Estimation. Instance segmentation is encoded as Center of Mass Regression, with performance improved by a positional-embedding (PE) based loss. It handles void regions and small instances using Edge Distance Sampling (EDS), focusing training on pixels near object boundaries. We are particularly interested in this approach because the self-supervised DinoV2 weights make it well-suited for handling both dense and sparse tasks with different loss optimizations.

%% file: sec/3_method.tex
\section{Method}
\label{sec:method}
\begin{figure*}[t]\centering
  % \hspace{-9px} % Adjust the value as needed 
  \includegraphics[width=.8\textwidth, height=0.55\linewidth]{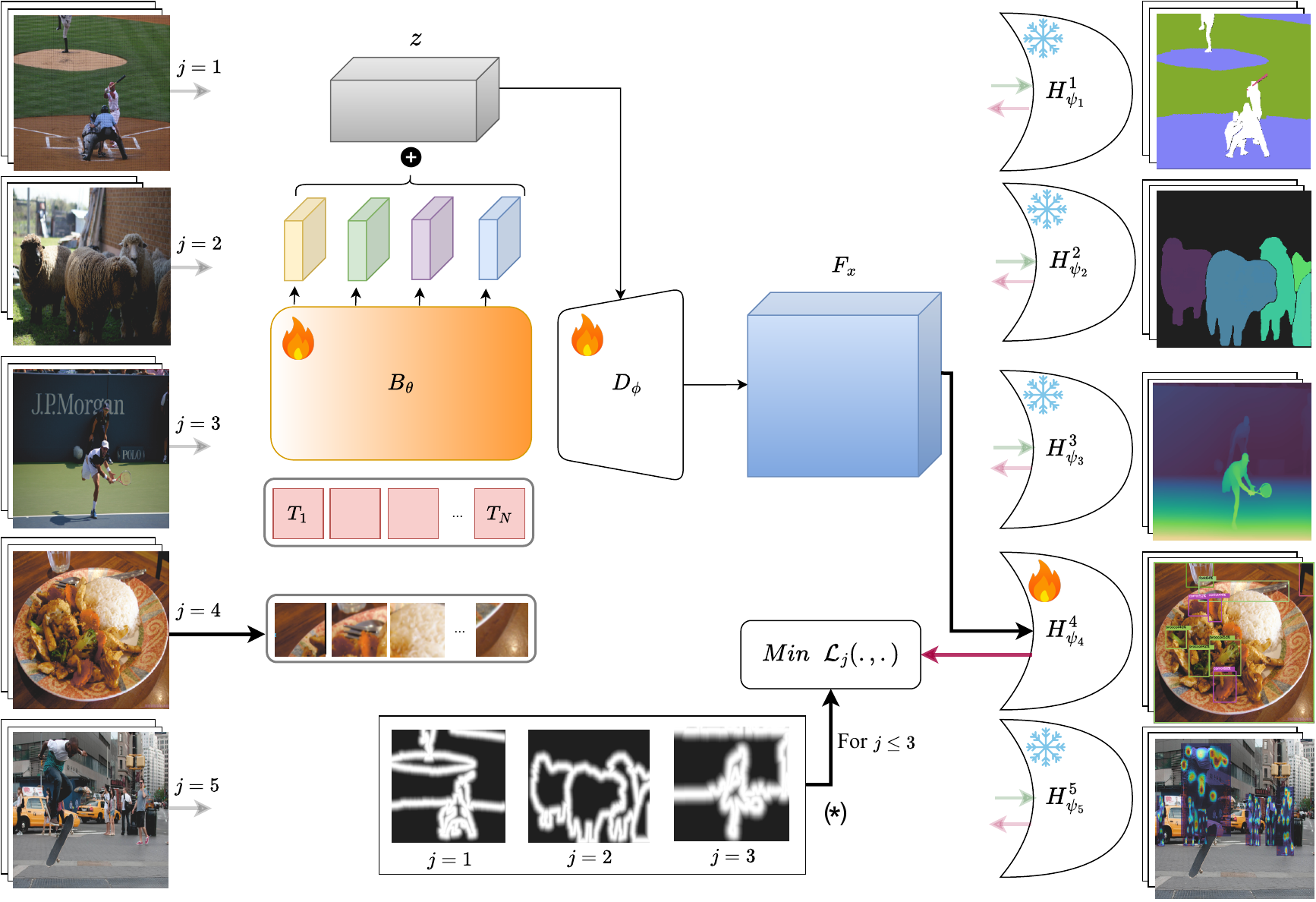}
   \caption{\textbf{Method Overview}. In each iteration, a random vision task—illustrated here with object detection—is selected. After patch tokenization, the backbone processes the tokens, and outputs from four intermediate layers are concatenated to form $z$. This is then decoded to a shared feature map $F_x=D_\phi(z)$, which is further processed by the corresponding task-specific projector. For the first three tasks, EDS is applied during loss minimization (*). The gradient-colored backbone highlights layer decay in $B_\theta$, where later layers receive stronger updates while earlier layers remain more stable, promoting better feature adaptation. Gray/dim arrows indicate inactive tasks, and for task projectors, input and outer arrows represent the forward and backward propagation, respectively.
}
   \label{fig:fig2}
   \vspace{-0.15in}
\end{figure*}

Given a set of task IDs \( T = \{1, 2, \dots, k\} \), each corresponding to a specific vision task, we collect datasets \( M_j \), where \( j \in T \), representing either individual datasets or a merged version of multiple datasets associated with task \( j \). We then establish a shared encoder backbone network \( {B}_{\theta} \) (referred to as the \emph{backbone} for simplicity) along with a relatively lightweight decoder \( D_{\phi} \), forming a feature extraction pipeline. Additionally, we define \( k \) shallow, task-specific projectors \( H^j_{\psi_j} \), each designed to align the output shape with the requirements of task \( j \).

Our objective is to train an encoder-decoder network that learns a shared and robust feature map \( F = D_{\phi}(B_{\theta}(\cdot)) \), ensuring it captures all the essential information needed to perform various vision tasks. This feature map \( F \) is then fed to task-specific projectors, whose main function is to adjust the output shape to match the requirements of each task, rather than extracting additional features from \( F \).

Specifically, we iteratively select a random vision task ID \( j \in T \) and sample a training data batch \( (x, y) \in M_j \), where \( x \) represents the input RGB data and \( y \) is the corresponding target. For simplicity, we define \( F_x = D_{\phi}(B_{\theta}(x)) \). In each iteration, the objective is to optimize the network by minimizing the task-specific loss function \( \mathcal{L}_j \left( H^j_{\psi_j} (F_x), y \right) \). Figure \ref{fig:fig2} illustrates the mechanism, including the network architecture and the training strategy.

\textbf{Backbone.} Our backbone is built upon ViT-L~\cite{dosovitskiy2020image}, utilizing weights obtained via the Dino-V2~\cite{oquab2023dinov2} strategy, which leverages self-supervised learning. This reliance on large-scale pretrained weights enables the model to learn rich, generalizable feature representations that are not tied to any specific task.
 This is particularly well-suited for our design, where task-specific projectors specialize the learned features for different tasks. This enables the model to efficiently share common features while adapting to the unique requirements of each task.

\textbf{Decoder.} The decoder comprises four transposed convolutional layers whose goal is to upscale the patch level representations to pixel level representations. Although the decoder is significantly lighter than the backbone, it is considerably larger than the task-specific projectors. This is because the task projectors are not intended to focus on deeply extracting new features, but to simply project this representation to the required output format for the task.

\subsection{Task-specific Projectors}
Given an RGB input $x$ of shape $H \times W \times 3$, Each intermediate layer of the backbone outputs a sequence of tokens with shape \( \left(\frac{H}{s} * \frac{W}{s}\right) \times \text{dim} \), where \( s \) is the patch size of the ViT network and \( \text{dim} \) is the dimension of the tokens. To construct a latent image representation \( z \), we concatenate tokens from four intermediate layers of the backbone. The upscaling decoder then processes this representation to generate a dense per-pixel feature map $F_x$ of shape $H \times W \times O$, where the channel dimension \( O \) is set to 256 in our experiments. Finally, task-specific projectors utilize \( F_x \) to optimize their respective loss functions based on the given task.

For the first three tasks (SS, IS, and DE) involving pixel-wise loss, we do not treat all pixels equally. Instead, we employ Edge Distance Sampling (EDS), which focuses training on pixels near object boundaries. For each pixel \((i, j)\), we calculate its Euclidean distance \(d_{ij}\) to the nearest boundary. The corresponding weight for that pixel is given by:
\[
w_{ij} = w_{\text{min}} + (1 - w_{\text{min}}) e^{-d_{ij}^2 / D^2},
\]
where \(w_{\text{min}}\) represents the minimum weight assigned to pixels far from boundaries. In our experiments, we set \(w_{\text{min}} = 0\). The parameter \(D\) controls how rapidly the weight diminishes with increasing distance from the boundary, with \(D = 20\) in our experiments.

Edge Distance Sampling addresses two critical issues:  
(i) \textit{Unlabeled regions} are effectively excluded, reducing ambiguity.  
(ii) \textit{Loss imbalance} between small and large objects is mitigated by concentrating on instance boundaries.

\textbf{SS Projector.} This projector formulates the semantic segmentation task (\(j=1\)) as per-pixel classification, following \cite{oquab2023dinov2}. 
Specifically, the loss function is a per-pixel weighted cross-entropy loss, formulated as:

\[
\mathcal{L}_1 \left( {H}^1_{\psi_1} ({F}_x):=\hat{y}, {y} \right) = \sum_{i=1}^{H} \sum_{j=1}^{W} w_{ij} \cdot CE \left( \hat{y}_{ij}, {y}_{ij} \right),
\]

where the cross-entropy loss \( CE(\cdot) \) is applied at each spatial location \((i, j)\) to refine segmentation accuracy. %In our experiments, 
The \(  {H}^1_{\psi_1} ({F}_x)  \) is a single \( 3 \times 3 \) convolution layer from channel dimension \( O \) to the number of categories in the dataset.

\textbf{IS Projector.}
We tackle class-agnostic instance segmentation (\(j=2\)) by encoding instances via their center of mass. Each pixel predicts the coordinates $(u, v)$ of the corresponding instance centroid, while unlabeled pixels receive a void encoding. The loss depends on the accuracy of centroid predictions, with higher precision needed in regions where centroids are close together (e.g., near object boundaries), as small errors can lead to incorrect instance assignments.

To address this, we apply positional embeddings to map centroid coordinates into a higher-dimensional space. The $u$ and $v$ coordinates are first normalized to the range $[-1, 1]$, then embedded as:
\[
\gamma(p) = \left[ \sin(2^k\pi p), \cos(2^k\pi p) \right]_{k=0}^{L-1} \in \mathbb{R}^{2L},
\]
where $L$ is the number of harmonics used, controlling the dimensionality of the embedding for coordinate $p$.

For a pixel $(i, j)$, we define a predictor function $\mathbb{P}(i, j)$ that returns a {concatenated positional embedding} $\gamma(u) \oplus \gamma(v)$ if pixel $(i, j)$ belong to an instance with centroid at $(u, v)$, and returns a zero vector $Z \in \mathbb{R}^{4L}$ if the pixel is unlabeled. Thus, the target pixel output \(y_{ij}\) is:

\[
y_{ij} = \mathbb{P}(i, j) \cdot \left( \gamma(u) \oplus \gamma(v) \right) + (1 - \mathbb{P}(i, j)) \cdot Z
\]
The {instance segmentation loss} is then defined as:
\[
\mathcal{L}_2 \left( {H}^2_{\psi_2} ({F}_x):=\hat{y}, {y} \right) = \sum_{i=1}^{H} \sum_{j=1}^{W} w_{ij} \cdot L(\hat{y}_{ij}, y_{ij}),
\]

where the pixel loss $L(\hat{y}_{ij}, y_{ij})$ is computed as the average euclidean distance between prediction and the target embeddings:
\[
L(\hat{y}_{ij}, y_{ij}) = \frac{1}{2}\sum_{n=1}^{2} \left\| ({y}_{ij}-\hat{y}_{ij} )_{[2(n-1)L+1:2nL]} \right\|^2
\]
In our experiments, \(  {H}^2_{\psi_2} ({F}_x)  \) is a single \( 3 \times 3 \) convolution layer from channel dimension \( O \) to $4L$.

During inference, the \( u \) and \( v \) coordinates are retrieved from \( \hat{y}_{ij} \) using the nearest neighbor search~\cite{prisadnikov2024simple}, matching predictions to positions in a discretized \( uv \)-grid space.

\textbf{Panoptic Segmentation}. We split panoptic segmentation into semantic and class-agnostic instance segmentation, solving them independently using the pipeline outlined in the SS and IS projectors sections.

\textbf{DE Projector.}
For monocular depth estimation (\( j = 3 \)), we minimize a weighted affine-invariant loss in disparity space. We define disparity \( d \) as the inverse depth, normalized between 0 and 1, with the depth range set from 0 to 10. The projection layer, \( {H}^4_{\psi_4} ({F}_x) \), consists of a single 3×3 convolution with a sigmoid activation. It maps the feature map from channel dimension \( O \) to a single-channel output and optimizes the following loss:

\[
\mathcal{L}_3 \left( {H}^3_{\psi_3} ({F}_x):=\hat{d}, {d} \right) = \sum_{i=1}^{H} \sum_{j=1}^{W} w_{ij} \cdot \left| \hat{d}_{ij} - d_{ij} \right|
\]

\textbf{OD Projector.}
For the object detection task (\( j = 4 \)), we follow a Yolo-style~\cite{redmon2016you,jiang2022review} approach but modify the architecture by replacing fully connected layers with convolutional layers for the final predictions. The last layer applies a sigmoid activation function to produce the output.

The image is divided into an \( S \times S \) grid, where each cell predicts \( B \) bounding boxes, each defined by \( (a, b, w, h, p) \). Here, \( a, b \) are the normalized center coordinates within the grid cell, \( w, h \) are normalized to the image, and \( p \) represents the predicted IoU with the ground truth.  

From the \( B \) predicted boxes per cell, we select the one with the highest IoU. Each grid cell $(i, j)$, also predicts a classification vector $\hat{c}_{ij}$ of length \( C \), where \( C \) is the number of object categories. The goal is for \( \hat{c}_{ij} \) to closely match the ground truth class vector \( c_{ij} \), which is a one-hot vector when an object is present in the grid cell and a zero vector otherwise. Thus,  \(  {H}^4_{\psi_4} ({F}_x)  \) produces an output of shape \( S \times S \times (5B + C) \).  

Specifically, the loss function consists of three main components: confidence loss (\( \mathcal{L}_{\text{p}} \)), coordinate loss (\( \mathcal{L}_{\text{abwh}} \)), and classification loss (\( \mathcal{L}_{\text{c}} \)):

\[
\mathcal{L}_{\text{p}} = \lambda_{{p}_{1}}\sum_{i,j} \hat{p}_{ij}^2 \cdot \mathbb{1}_{\text{noobj}} + 
\lambda_{{p}_{2}}\sum_{i,j} \left( \hat{p}_{ij} - \hat{\text{IoU}}_{ij} \right)^2 \cdot \mathbb{1}_{\text{obj}},
\]
\[
\begin{aligned}
\mathcal{L}_{\text{abwh}} = \lambda_{a} \sum_{i,j}  \Big(& \| \hat{a}_{ij} - a_{ij} \|^2 + \| \hat{b}_{ij} - b_{ij} \|^2 + \\
& \| \hat{w}_{ij}^{\frac{1}{2}} - w_{ij}^{\frac{1}{2}} \|^2 + \| \hat{h}_{ij}^{\frac{1}{2}} - h_{ij}^{\frac{1}{2}} \|^2 \Big) \cdot \mathbb{1}_{\text{obj}},
\end{aligned}
\]

\[
\begin{aligned}
\mathcal{L}_{\text{c}} = \lambda_{c}\sum_{i,j} \left( \hat{c}_{ij} - c_{ij} \right)^2 \cdot \mathbb{1}_{\text{obj}}, 
\end{aligned}
\]
where \( \mathbb{1}_{\text{obj}} \) is the boolean mask that selects the grid cells containing objects, \( \mathbb{1}_{\text{noobj}} \) is the boolean mask for grid cells without objects, and $\cdot$ performs tensor masking. The \( \hat{\text{IoU}} \) represents the predicted IoU between the best predicted bounding box and the ground truth bounding box during training. For training, we set the weighting coefficients as $10\lambda_{p_1}=2\lambda_{p_2}=\lambda_{a}=5\lambda_{c}$, leading to the following loss:
\[
\mathcal{L}_4 \left( {H}^4_{\psi_4} ({F}_x), {y} \right) = \mathcal{L}_{\text{p}} + \mathcal{L}_{\text{abwh}} +
\mathcal{L}_{\text{c}}
\]
During inference, we utilize class-aware and DIoU-based Non-Maximum Suppression (NMS) to effectively eliminate duplicate bounding boxes.

\textbf{KD Projector.}
We solve the 17-joint keypoint detection problem \((j=5)\) using a top-down approach, where cropped instances of individuals are processed independently. Instead of predicting separate heatmaps for each joint, we generate a single unified heatmap capturing all joints simultaneously, along with 17 classification channels trained exclusively on ground truth keypoints. We found this formulation significantly more efficient than the conventional approach of predicting one heatmap per joint. With all but one channel representing discrete class labels, this encoding is particularly well-suited for knowledge distillation, enabling the proposal of KD\textsuperscript{*} (\ref{sec:KD_star}). The projector \(  {H}^5_{\psi_5} ({F}_x)  \) is a single \( 3 \times 3 \) convolution layer from channel dimension \( O \) to 18.

To optimize the model, we employ a hybrid loss function combining per-pixel mean squared error (MSE) loss for heatmap regression and cross-entropy loss for joint classification, applied only at ground truth locations:
\[
\mathcal{L}_5 \left( {H}^5_{\psi_5} ({F}_x), {y} \right) = \text{MSE}\left(\hat{ht}, ht^{\sigma}\right) + \sum_{i,j} \text{CE} \left( \hat{c}_{ij}, c_{ij} \right) \cdot \mathbb{1}_{\text{gt}}
\]

The last channel of ${H}^5_{\psi_5} ({F}_x)$, \( \hat{ht} \), is the predicted heatmap of size $hs$, and \( ht^{\sigma} \) represents a Gaussian distribution palette centered at each keypoint location with standard deviation \( \sigma \). Additionally, \( \hat{c} \), the first 17 channels of ${H}^5_{\psi_5} ({F}_x)$, is the predicted class values. \( c_{ij} \) is the one-hot ground truth class vector at heatmap location \( (i,j) \), and \( \mathbb{1}_{\text{gt}} \) is a binary mask selecting the pixels that correspond to ground truth class labels.

During inference, we apply a class-aware argmax operation to the heatmap, selecting the highest-scoring location for each joint while filtering out low-confidence detections.

\begin{figure}[t]
  \hspace{-10px} % Adjust the value as needed 
  \includegraphics[width=0.5\textwidth, height=0.55\linewidth]{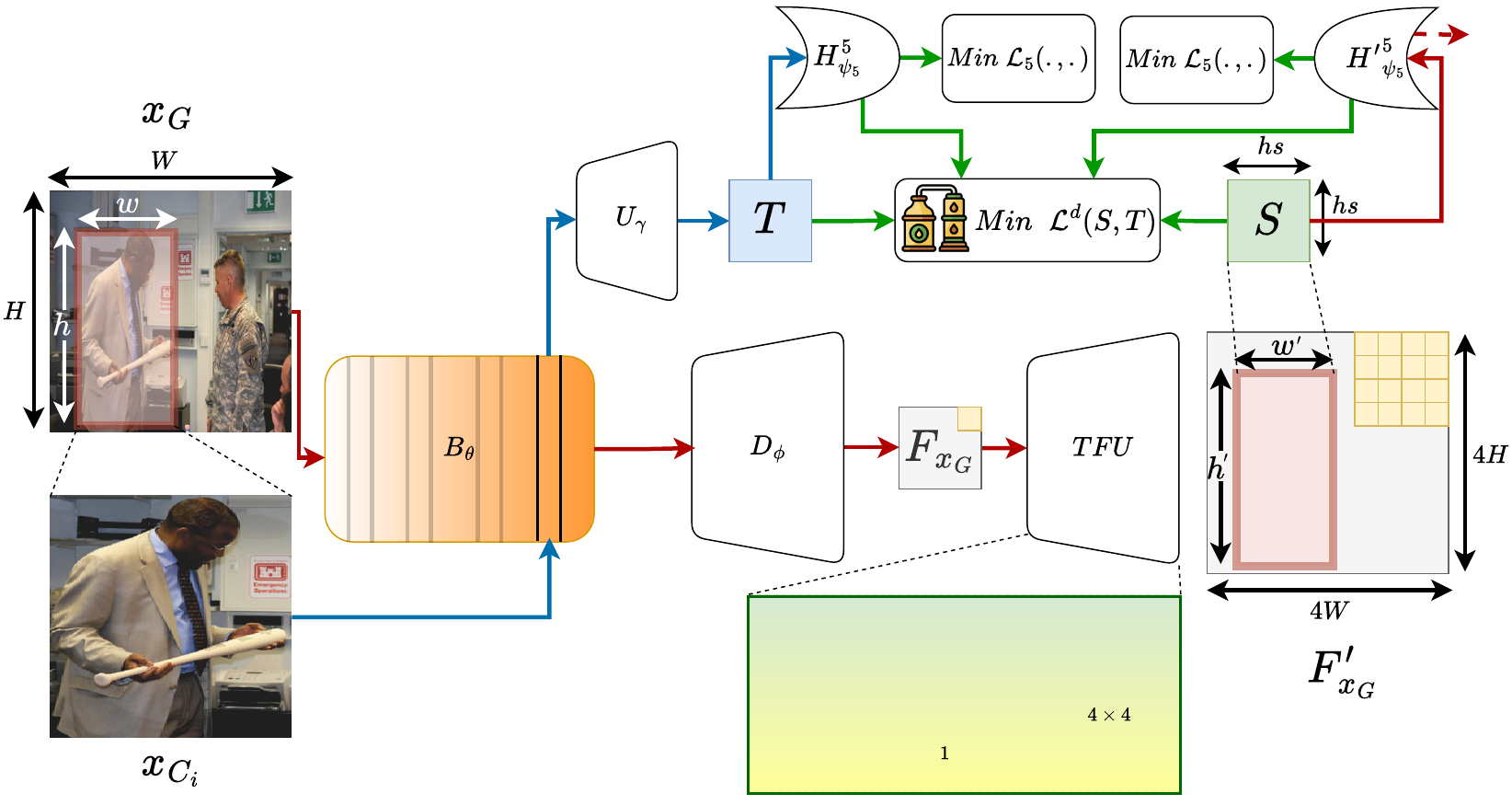}
   \caption{\textbf{KD\textsuperscript{*} Overview.} Blue arrows indicate paths that are inactive during inference, while red arrows represent the active inference path. By enabling feature cropping, we reduce the number of forward passes from \( N \) to 1. In this process, ${}_4z$ represents the latent image embedding obtained from the last of the four intermediate layers, \( w' = 4w \), and \( h' = 4h \).  The green lines indicate contributions to the loss function, while the dashed red line represents the point at which we obtain the final output during the inference stage.
}
\vspace{-0.1in}
   \label{fig:fig3}
\end{figure}
\subsection{KD\textsuperscript{*}:Single-pass Top-down Keypoint Detection}\label{sec:KD_star}
In top-down keypoint detection, the entire network must run repeatedly for each instance in an image, leading to increased inference time and resource consumption—an issue exacerbated by the use of a large backbone. This contradicts the essence of efficient multitasking. %, where multiple tasks should be performed in a single forward pass. 
To overcome this, we leverage knowledge distillation to enable the keypoint detection with just one forward pass for the entire image.

In our streamlined implementation of the keypoint detection task, we found that strong optimization performance could be achieved by directly using a part of the latent image representation \( {}_4z = {}_4B_\theta(x) \), bypassing the decoder (indicated by the blue arrows in Figure \ref{fig:fig3}). This approach generates a feature map suitable for cropped RGB images of individuals, but in a much more efficient manner, without being part of the multitasking pipeline. Building on this observation, we propose a knowledge distillation paradigm that aligns the feature map of cropped images with the corresponding regions of the full-image feature map. This technique allows us to reduce the forward passes from one per individual to a single pass for the entire image, essentially cropping the feature map instead of the RGB input.

Let \( x_G \) be an image containing \( N \) individuals, and \( x^i_C \) be the cropped image corresponding to the \( i \)-th person. We define a cropping function \( Q(F, \mathrm{Box}) \) that returns the cropped portion of the feature map \( F \) based on the bounding box \( \mathrm{Box} \), interpolated to the same size as \( F \). It is easy to see that \( Q(x_G, \mathrm{Box}_i) = x^i_C \). The goal is to minimize the disparity between student \( S = Q(F'_{x_G}, \mathrm{Box}_i) \) and teacher \( T = U_\gamma({}_4B_\theta(Q(x_G, \mathrm{Box}_i))) \), as shown in Figure \ref{fig:fig3}, where \( U_\gamma \) is a shallow upscaler. In the inference stage, we replace the \( N \) forward runs of \( D_{\phi}(B_{\theta}(Q(x_G, \mathrm{Box}_i)))_{i=1}^{N} \) with \( N \) cropping operations \( Q(D_{\phi}(B_{\theta}(x_G)), \mathrm{Box}_i)_{i=1}^{N} \), reducing the computation to a single forward run.

A challenge with our feature-cropping strategy in \( Q(F_{x_G}, \mathrm{Box}_i) \) is that small boxes result in a significant loss of spatial information in \( F_{x_G} \), leading to suboptimal features. To address this, we introduce a training-free upscaler \( TFU \) that increases the spatial resolution of \( F_{x_G} \) by a factor of 4, enhancing its robustness to small boxes. The resulted feature map is denoted as \( F'_{x_G} \). More specifically, \( TFU \) processes an \( H \times W \times O \) feature map by dividing the \( O \) channels into \( O/16 \) subchannels, each containing a \( 4 \times 4 \) grid of values. The grid then replaces the original pixel at each spatial location, enhancing the spatial resolution by 4x while reducing the number of channels by \( 1/16 \)x.

We use a hybrid loss to distill information from $T$ to $S$:
\[
\begin{aligned}
\mathcal{L}^*(S, T) &= \underbrace{\mathcal{L}_5 \left( \hat{y}_t, {y} \right)}_{\mathcal{L}_1^*} + \underbrace{\mathcal{L}_5 \left( \hat{y}_s, {y} \right)}_{\mathcal{L}_2^*} + \mathcal{L}^d \left( S,T \right),
\end{aligned}
\]

with the distillation loss function defined as: 
\[
 \mathcal{L}^d \left( S,T \right) = \underbrace{\rho_1 \| S - T \|_1}_{\mathcal{L}^d_1} + \underbrace{\rho_2 \text{KL}({\hat{y}_t}^{:17} \parallel {\hat{y}_s}^{:17})}_{\mathcal{L}^d_2},
\]

where \( \hat{y}_t = {H}^5_{\psi_5} (T) \) and \( \hat{y}_s = {H'}^5_{\psi_5} (S) \). Note that \( {H}^5_{\psi_5} \) and \( {H'}^5_{\psi_5} \) have identical network architectures, but do not share weights. KL denotes the Kullback-Leibler divergence loss, and the superscript \( ^{:17} \) indicates the first 17 channels of the corresponding tensor. In our experiments, we set \( \rho_1 = 1.0 \) and \( \rho = 5 \times 10^{-5} \).

%% file: sec/4_exps.tex
\section{Experiments}
\label{sec:exp}

\subsection{Setup and Implementation}
% \textbf{Training Details.}
\textbf{Three-Stage Training.}
We adopt a three-stage training strategy. In the first stage, we train only on the IS task, the most challenging one, for a substantial number of epochs using low-resolution images (280$\times$280). In the second stage, we extend training to all tasks while maintaining the same low resolution and training for many epochs. Finally, in the third stage, we fine-tune the pretrained weights on high-resolution images (616$\times$616) with significantly reduced augmentation and for considerably fewer epochs. The idea here is that most of the training occurs on computationally efficient low-resolution data, while fine-tuning on high-resolution images refines the model effectively without incurring excessive computational cost.

\fyq{The three training stages run for 280, 150, and 60 epochs, respectively, totaling 5 days. Each stage uses a $5 \times 10^{-4}$ learning rate, 0.6 decay, a $1 \times 10^{-6}$ minimum learning rate, and 0.6 layer decay for the backbone.}
% The first, second, and third stages are trained for 280, 150, and 60 epochs, respectively, for a total combined training time of 5 days, with each stage using a learning rate of $5 \times 10^{-4}$, a learning rate decay of 0.6, a minimum learning rate of $1 \times 10^{-6}$, and a layer decay of 0.6 for training the backbone. 
A warmup period of 10 epochs is applied. The effective batch sizes are 1024, 512, and 256 for the three stages, respectively. We use 4 intermediate layers of the backbone, specifically layers 4, 11, 17, and 23. We set $L = 4$ in IS task, as ablated in Table~\ref{tab:tab3}. In the object detection task, we use $(B, S)$ of $(2, 7)$ and $(2, 10)$ in the second and third stages, respectively. For KD task, the heatmap size, $hs$, is set to $4/7$ ($\sigma=10$) of the image resolution in the second and third stages. For tasks involving EDS, we include an additional total variation loss as a regularizer. Specifically for the IS task, we also incorporate generalized DICE loss (SupMat 1.2).

\textbf{More Training Details.}
Training is conducted on 8$\times$A100 GPUs, with the ViT-Large backbone $B_{\theta}$ and CNN-based upscaler decoder $D_{\phi}$ containing 305M and 4M parameters, respectively. We target the COCO benchmark for training all five tasks. For depth estimation, we use pseudo-labels from DepthAnythingV2~\cite{yang2024depth,yang2025depth}. Experiments on $KD^{*}$ are conducted only at a low resolution of 280 within the previously stated training setup, using $U_{\gamma}$ as a 1M-parameter upscaler, constructed by 3 transposed convolutional layers.

\begin{table*}[h]
    \centering
    \renewcommand{\arraystretch}{1.3} % Adjust vertical spacing
    \resizebox{\textwidth}{!}{%
    \begin{tabular}{c|c|c|c|cc|c|cc|cc}
        \hline
        \textbf{Model} & \textbf{Resolution} & \textbf{\#Params} & \textbf{Backbone} & \multicolumn{2}{c|}{\textbf{Segmentation}} & \textbf{Depth} & \multicolumn{2}{c|}{\textbf{Detection}} & \multicolumn{2}{c}{\textbf{Keypoint}} \\
        \cline{5-6} \cline{7-7} \cline{8-9} \cline{10-11}
        & & & & \textbf{PQ $\uparrow$} & \textbf{mIoU $\uparrow$} & \textbf{RMSE $\downarrow$} & \textbf{mAP\textsubscript{50} $\uparrow$} & \textbf{mAP\textsubscript{all} $\uparrow$} & \textbf{OKS-mAP\textsubscript{50} $\uparrow$} & \textbf{OKS-mAP\textsubscript{all} $\uparrow$} \\
        \hline
        \multicolumn{11}{c}{\cellcolor{gray!20} \textbf{Specialized Models}} \\
        \hline
        MaskFormer & $640\times640$ & 212M & SWIN-L & 52.7 & - & $\times$ & $\times$ & $\times$ & $\times$ & $\times$ \\
        \hline
        Mask2Former & $640\times640$ & 216M & SWIN-L & 57.8 & - & $\times$ & $\times$ & $\times$ & $\times$ & $\times$ \\
        \hline
        OneFormer & $640\times640$ & 223M & DINAT-L & \textbf{58.0} & \textbf{67.4} & $\times$ & $\times$ & $\times$ & $\times$ & $\times$ \\
        \hline
        Faster-RCNN~\cite{girshick2015fast} & $600\times1000$ & 60M & ResNet-101 & $\times$ & $\times$  & $\times$  & - & 44.0 & $\times$  & $\times$  \\
        \hline
        DETR~\cite{carion2020end} & $800\times800$ & 60M & ResNet-101/DC5  & 45.1 & - & $\times$  & \textbf{64.7} & \textbf{44.9} & $\times$  & $\times$  \\
        \hline
        HRNET~\cite{wang2020deep} & $256\times192$ & 28.5M & HRNET-V1-W32 & $\times$  & $\times$  & $\times$  & $\times$  & $\times$  & \textbf{90.5} & \textbf{74.4} \\
        \hline
        \multicolumn{11}{c}{\cellcolor{gray!20} \textbf{Multi-Task Models}} \\
        \hline
        LDM~\cite{van2024simple} & $512\times512$ & 850M & UNET & 43.3 & 60.1 & 0.075\textsuperscript{*} & $\times$  & $\times$  & $\times$  & $\times$  \\
        \hline
        Pix2Seq-D & $1024\times1024$ & 94.5M & ResNet-50 & 50.3 & - & $\times$  & $\times$  & $\times$  & $\times$  & $\times$  \\
        \hline
        Pix2Seq & $1024\times1024$ & 132M & ViT-B & $\times$  & $\times$  & $\times$  & - & 46.5 & - & 64.8 \\
        \hline
        Mask-RCNN~\cite{wang2018non} & $320\times320$ & 43M & ResNet-101+FPN & $\times$  & $\times$  & $\times$  & 67.8 & 45.0 & 87.3 & 66.5 \\
        \hline
        UViM~\cite{kolesnikov2022uvim} & $1280\times1280$ & 939M & ViT-L & 45.8 & - & - & $\times$  & $\times$  & $\times$  & $\times$  \\
        \hline
        Painter & $448\times448$ & 303.5M & ViT-L & 43.4 & - & - & $\times$  & $\times$ & - & \textbf{72.1} \\
        \hline
        GiT\textsuperscript{\textbf{\dag}} & $1120^2|672^2$ & 756M & Multi-layer Transformer & $\times$  & 52.4 & $\times$  & \textbf{71.0} & \textbf{52.9} & $\times$  & $\times$  \\
        \hline
        {Unified-IO\textsuperscript{\textbf{\dag}}\textsubscript{XL}} & $384\times384$ & 2.9B & ViT-L & $\times$ & 56.5 & 0.385\textsuperscript{*} & $\times$ & $\times$ & - & 68.1 \\
        \hline
        AHMAD\textsubscript{single} & $616\times616$ & 309.5M & ViT-L & {55.1} & {66.9} & {0.331} & {64.3} & {43.9} & {90.8} & {67.3} \\
        \hline
        \rowcolor{yellow!20} AHMAD\textsubscript{multi} & $616\times616$ & 309.5M & ViT-L & \textbf{53.1} & \textbf{66.5} & \textbf{0.310} & {62.9} & {44.6} & \textbf{92.2} & {68.2} \\
        \hline
    \end{tabular}%
    }
    \caption{\textbf{Comparison of multi-task models across various evaluation metrics on the COCO-val split.} GiT performs object detection and instance segmentation at a resolution of $1120\times1120$, while semantic segmentation is conducted at $672\times672$. Additionally, Unified-IO is evaluated on the GRIT benchmark, which includes the COCO dataset. "$\times$" indicates that the model does not support the task, while "-" means the task is supported, but no performance numbers for the COCO dataset are provided in the paper.}
    \label{tab:tab1}
\end{table*}

\subsection{Main Results on COCO}
% \textbf{Evaluation.}
We evaluate the performance of our multi-task learning pipeline using task-specific metrics on the COCO-val split. Specifically, for panoptic segmentation, we report Panoptic Quality (PQ), and for semantic segmentation, we use mean Intersection over Union (mIoU). For depth estimation, we report Root Mean Squared Error (RMSE), for object detection, we use mean Average Precision (mAP), and for keypoint detection, we provide OKS-based mAP. Table \ref{tab:tab1} compares these results with those from other specialized and multi-task generalist vision models, highlighting key performance trends.

Notably, our approach achieves strong performance in panoptic segmentation, surpassing other models with a PQ of 53.1 and an mIoU of 66.5. This represents a 22.3\% boost compared to Painter and a 5.5\% boost compared to Pix2Seq-D. As for semantic segmentation, we outperform LDM by a \( \Delta \text{mIoU} \) of 6.4 and significantly surpass GiT with a \( \Delta \text{mIoU} \) of 14.1. For depth estimation, we only compare the results of AHMAD\textsubscript{single} to AHMAD\textsubscript{multi} (since other models use different ground truth labels for training, as noted by * in Table \ref{tab:tab1}), highlighting the complementary boost gained from joint training. We also achieve competitive results with an OKS-mAP of 68.12 for keypoint detection and an mAP of 44.6 for object detection, demonstrating the effectiveness of our multi-task framework in improving performance through joint training. 

\begin{figure}[h!]
   \vspace{-0.2in}\centering
  % \hspace{-19px} % Adjust the value as needed 
  \includegraphics[width=0.45\textwidth, height=1.15\linewidth]{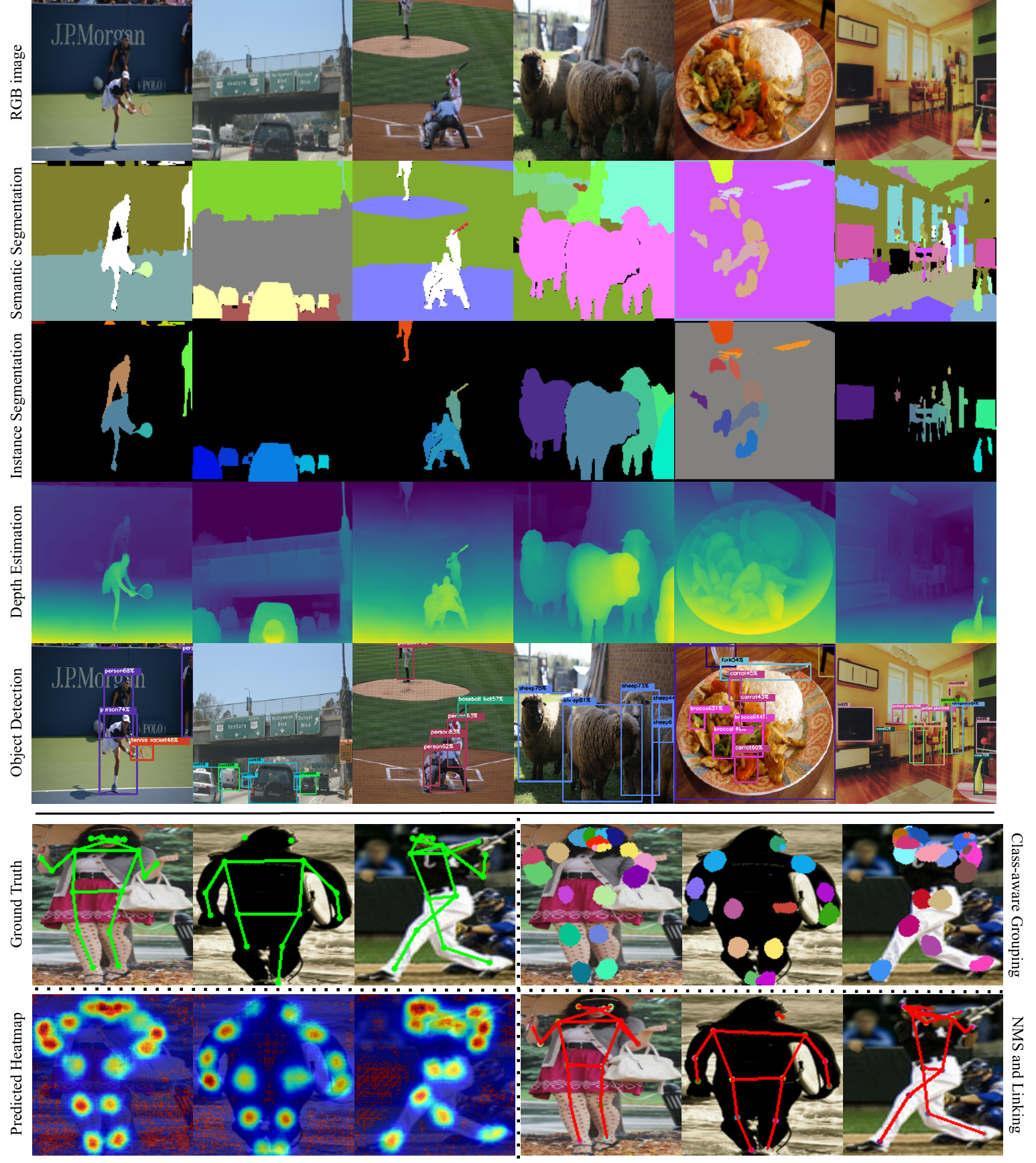}
     \vspace{-0.1in}
   \caption{\textbf{Qualitative Results.} \fyq{We show our results on five different tasks with diverse examples.}}
   \label{fig:fig4}
   \vspace{-0.2in}
\end{figure}

% For qualitative results, see the example predictions in Figure \ref{fig:fig4}.
\fyq{Figure \ref{fig:fig4} provides a visualization of our approach. Several examples with predictions on segmentation masks, depth estimations, object detections, and keypoint predictions, demonstrating how our model generalizes across different tasks within a unified framework.}

As for our proposed KD\textsuperscript{*} method, we demonstrate how distilling information from the teacher feature map to the student feature map allows us to reduce the number of forward passes by a factor of 2.7x. This makes the entire multi-task paradigm unified, where the model processes the same input image for different tasks and generates the results in a single forward pass. Table \ref{tab:tab2} presents the statistics. The results from Table \ref{tab:tab4} further demonstrate that, with sufficient engineering effort, performance progressively improves, validating the efficacy of the approach. Additionally, there is potential for further enhancement as more engineering is applied.

\begin{table}[h]
    \centering
    \renewcommand{\arraystretch}{1.75} % Adjust vertical spacing
    \resizebox{\columnwidth}{!}{%
    \begin{tabular}{c|c|c|c|c|c}
        \hline
        \textbf{Method} & \textbf{Input} & \textbf{\#Projector Params} & \textbf{OKS-mAP $\uparrow$} & \textbf{Avg Run/Img} & \textbf{Decoder} \\
        \hline
        Top-down KD & $C_{280\times 280}$ & 440K & 62.6 & 2.7 & \checkmark \\
        \hline
        Bypassed KD & $C_{280\times 280}$ & 2.5K & 60.1 & 2.7 & $\times$ \\
        \hline
        \rowcolor{yellow!20} KnowDist KD & $G_{280\times 280}$ & \textbf{2.5K} & \textbf{41.32} & \textbf{1.0} & \checkmark \\
        \hline
    \end{tabular}%
    }
    \caption{\textbf{A comparative analysis} of the standard top-down Keypoint detection, the decoder-bypassed variant, and our proposed Knowledge Distillation approach. $G$ denotes the global image input, while $C$ represents the cropped image input. The last column shows a $\checkmark$ for methods that run the decoder, and $\times$ otherwise.}
    \label{tab:tab2}
\end{table}

Beyond the performance gains achieved through knowledge distillation, it is noteworthy that the bypassed path alone achieves strong metric scores. This highlights the suitability and effectiveness of the DINO-V2 backbone within our hybrid training framework. To better approximate real-world inference in top-down keypoint detection, we introduce 20\% noise into the ground-truth bounding boxes when cropping input images.

% \textbf{Ablation Study.}
\section{Ablation Study}
For semantic segmentation, we investigated the impact of the EDS mechanism. In the case of instance segmentation, we also explored the effect of positional embeddings across different values of \( L \), as summarized in Table \ref{tab:tab3}.

\begin{table}[h]
    \centering
    \renewcommand{\arraystretch}{1.0} 
    \begin{tabular}{c|c|c}
        \hline
        \multicolumn{3}{c}{\textbf{Panoptic Segmentation}} \\
        \hline
        EDS & (u,v) Encoding & PQ $\uparrow$ \\
        \hline
        $\times$ & 2D Regression & 38.01 \\
        $\times$ & PE L=4 & 44.86 \\
        $\checkmark$ & 2D Regression & 41.61 \\
        $\checkmark$ & Cross-Entropy & 39.87 \\
        $\checkmark$ & PE L=3 & 47.38 \\
        $\checkmark$ & PE L=4 & \textbf{48.61} \\
        $\checkmark$ & PE L=5 & 48.12 \\
        \hline
    \end{tabular}
    \caption{\textbf{Panoptic Segmentation Ablation.} Effects of EDS and Positional Embedding (PE). In 2D Regression, we directly regressed $(u, v)$. The ablation is performed on a ViT-Base model with a resolution of $420\times420$.}
    \label{tab:tab3}
\end{table}

For keypoint and object detection, we primarily conducted ablation studies on various hyperparameters, such as $\sigma$ in keypoint detection and the number of grids \( S \) in object detection, along with different NMS versions for post-processing. For example, we compared using DBSCAN and KMeans to obtain the final joints from candidates in keypoint detection, as well as IoU/dIoU-based NMS in object detection. In our post-processing analysis, we found that joint-aware argmax over heatmap values outperforms clustering-based methods, and dIoU-based NMS performs better than IoU-based NMS in object detection.

Finally, for the Knowledge Distillation approach, we primarily examined the effect of various loss functions and different training strategies, as detailed in Table \ref{tab:tab4}.
\begin{table}[h]
    \centering
    \renewcommand{\arraystretch}{1.5} % Adjust vertical spacing
    \resizebox{\columnwidth}{!}{%
    \begin{tabular}{c|c|c|c}
        \hline
        \textbf{Conf ID} & \textbf{Config} & \textbf{Loss Function} & \textbf{OKS-mAP} $\uparrow$ \\
        \hline
        CID-1 & Stand-alone feature cropping & $\mathcal{L}^*_2$ & 34.2 \\
        \hline
        CID-2 & CID-1, RGB concatenation & $\mathcal{L}^*_2$ & 34.6 \\
        \hline
        CID-3 & Distillation, Shared Projector & $\mathcal{L}^*_1+\mathcal{L}^d_1$ & 35.2 \\
        \hline
        CID-4 & CID-3 & $\mathcal{L}^*_1+\mathcal{L}^d_1+\mathcal{L}^*_2$ & 35.6 \\
        \hline
        CID-5 & CID-4, Different Projectors & $\mathcal{L}^*_1+\mathcal{L}^d_1+\mathcal{L}^*_2$ & 36.2 \\
        \hline
        CID-6 & CID-5, TFU added & $\mathcal{L}^*_1+\mathcal{L}^d_1+\mathcal{L}^*_2$ & 39.8 \\
        \hline
        CID-7 & CID-6 & $\mathcal{L}^*_1+\mathcal{L}^d_1+\mathcal{L}^*_2+\mathcal{L}^d_2$ & \textbf{41.3} \\
        \hline
    \end{tabular}%
    }
    \caption{\textbf{KD\textsuperscript{*} Ablation.} Comparison of different architectures and loss functions on OKS-mAP performance. For CID-1 and CID-2, no distillation is performed. In CID-2, the cropped image is directly concatenated with the feature map from the input. In CID-3, both the student and teacher utilize the same projector, \( {H}^5_{\psi_5} \). Conversely, in CID-5, the student and teacher employ different but architecturally identical projectors, \( {H}^5_{\psi_5} \) and \( {H'}^5_{\psi_5} \). Observe how our TFU function effectively enhances performance.}
    \label{tab:tab4}
\end{table}

%% file: sec/5_conc.tex
\section{Conclusion}
\label{sec:conc}
\fyq{In this paper, we introduce AHMAD, a generalist multitasking framework that integrates five key vision tasks—semantic segmentation, instance segmentation, depth estimation, keypoint detection, and object detection—within a unified model. Our approach efficiently handles heterogeneous task outputs without complex serialization or pre/post-processing. To further improve efficiency, we propose a knowledge distillation strategy for keypoint detection, enabling a single forward pass instead of multiple passes required by traditional top-down methods. Experiments demonstrate strong performance across all tasks, with state-of-the-art results in panoptic and semantic segmentation.  AHMAD offers a simple yet effective multitask framework, balancing efficiency and flexibility for real-world applications. Future work includes refining distillation techniques, incorporating higher-resolution training, and expanding to additional vision tasks.}

\section{Acknowledgements}
This research was partially funded by the Ministry of Education and Science of Bulgaria (support for INSAIT, part of
the Bulgarian National Roadmap for Research Infrastructure).